%% file: Maree_UHVGOMEA.tex
\documentclass[twoside]{article}
\usepackage{ecj,palatino,epsfig,latexsym,natbib}
\usepackage[utf8x]{inputenc}
\usepackage{amsfonts}
\usepackage{amsmath}
\usepackage{amssymb}

\include{math_shortcuts}

\usepackage{multirow,tabularx}
\graphicspath{{images/}}

\DeclareMathSymbol{\dv}{\mathbin}{operators}{"3A}

\usepackage{graphicx}
\usepackage{multirow,tabularx}
\usepackage[hidelinks]{hyperref}
\usepackage{xcolor}

\usepackage{booktabs} 

\RequirePackage[ruled,vlined]{algorithm2e}
\SetAlFnt{\smaller}

\begin{document}

\ecjHeader{x}{x}{xxx-xxx}{201X}{UHV-based MO optimization with GOM}{S.C.Maree et al.}
\title{\bf Uncrowded Hypervolume-based Multi-objective Optimization with Gene-pool Optimal Mixing}

\author{\name{\bf S.C. Maree} \hfill \addr{s.c.maree@cwi.nl}\\ 
        \addr{Amsterdam UMC, University of Amsterdam, The Netherlands}\\
        \addr{Centrum Wiskunde \& Informatica, Amsterdam, The Netherlands}
\AND
       \name{\bf T. Alderliesten} \hfill \addr{t.alderliesten@lumc.nl}\\
        \addr{Leiden University Medical Center, The Netherlands}
\AND
\name{\bf P.A.N. Bosman} \hfill \addr{peter.bosman@cwi.nl}\\ 
        \addr{Centrum Wiskunde \& Informatica, Amsterdam, The Netherlands}
}

\maketitle

\begin{abstract}
Domination-based multi-objective (MO) evolutionary algorithms (EAs) are today arguably the most frequently used type of MOEA. These methods however stagnate when the majority of the population becomes non-dominated, preventing convergence to the Pareto set. Hypervolume-based MO optimization has shown promising results to overcome this. Direct use of the hypervolume however results in no selection pressure for dominated solutions. The recently introduced Sofomore framework overcomes this by solving multiple interleaved single-objective dynamic problems that iteratively improve a single approximation set, based on the uncrowded hypervolume improvement (UHVI). It thereby however loses many advantages of population-based MO optimization, such as handling multimodality. Here, we reformulate the UHVI as a quality measure for approximation sets, called the uncrowded hypervolume (UHV), which can be used to directly solve MO optimization problems with a single-objective optimizer. We use the state-of-the-art gene-pool optimal mixing evolutionary algorithm (GOMEA) that is capable of efficiently exploiting the intrinsically available grey-box properties of this problem. The resulting algorithm, UHV-GOMEA, is compared to Sofomore equipped with GOMEA, and the domination-based MO-GOMEA. In doing so, we investigate in which scenarios either domination-based or hypervolume-based methods are preferred. Finally, we construct a simple hybrid approach that combines MO-GOMEA with UHV-GOMEA and outperforms both.
\end{abstract}


\section{Introduction}
A multi-objective (MO) optimization problem is given by a to-be-minimized objective function $\bff : \cX \rightarrow \bbR^m$, with $\bff = [f_1,\ldots,f_m]$ being an $m$-dimensional vector function and $\cX \subseteq \bbR^n$ the $n$-dimensional decision space.  MO problems do not naturally imply a complete ordering of solutions $\bx\inX$ based on their objective values. When the objectives contradict each other, no single solution minimizes all objectives simultaneously. The optimum of an MO problem can then be defined in terms of \textit{Pareto optimality} \citep{Knowles2006}. A solution is said to \textit{dominate} another solution when it is strictly better in one or more objectives, and is not worse in any of the other objectives. A solution is \textit{Pareto optimal} when there exist no solutions that dominates is. The \textit{Pareto set} is the set of all Pareto optimal solutions, and its image under $\bff$ is known as the \textit{Pareto front}. In practice, the aim of MO optimization is to provide a decision maker with a set of non-dominated solutions, known as an \textit{approximation set} $\cA\subset\cX$, whose image under $\bff$, the \textit{approximation front}, approximates the Pareto front. The decision maker then selects a preferred solution from this set. Real-valued MO problems generally have an infinite number of Pareto-optimal solutions. As it is impossible to obtain all of these, MO optimization algorithms generally attempt to find an approximation set containing a manageable number of solutions that forms a good representation of the entire front. This results in an inherent trade-off in the two-sided optimization goal of MO optimization, since it is desirable to obtain a diverse approximation set as well as an approximation set that is close to the Pareto set \citep{bosman03}. 

MO evolutionary algorithms (MOEAs), have shown to be very successful for (black-box) MO optimization in practice \citep{Deb01book}. These algorithms maintain a population of solutions, and generally equip a domination-based fitness selection scheme \citep{deb02,zitzler01,bouter17b}, such as in the non-dominated sorting genetic algorithm (NSGA-II) \citep{deb02}, arguably the best-known MOEA. However, if the population size is (much) smaller than the number of Pareto-optimal solutions, at some point in the optimization process the majority of solutions in the population will be non-dominated. MOEAs then typically aim to improve diversity in the population, for example using the crowding distance in NSGA-II. This could lead to cyclic behavior, with the MOEA improving diversity but worsening proximity. Consequently, even though the approximation sets obtained with domination-based MOEAs are often sufficient for practical use, these algorithms do not result in approximation sets converging to the Pareto set \citep{Berghammer12}. Indicator-based MO optimization, especially based on the hypervolume measure \citep{Zitzler99}, has shown promising results to overcome this limitation \citep{Beume07,Igel07}. 

Practically used indicators in indicator-based optimization are the R2 indicator \citep{Hansen98}, the epsilon indicator \citep{Zitzler04}, and the hypervolume measure \citep{Zitzler2003}. The hypervolume measure is particularly interesting as it is the only known measure that is strictly monotonic with respect to Pareto-dominance \citep{Knowles2002,Fleischer2003}. This means that the approximation set with optimal hypervolume is a subset of the Pareto set. A limitation of the hypervolume measure is that dominated solutions have no contribution, in the sense that there is no selection pressure for dominated solutions towards a non-dominated region in the decision space \citep{Emmerich07}. Therefore, in the first usages of the hypervolume measure in the optimization process, it was only partially used to guide the search \citep{Emmerich05,Zitzler07,Igel07,Nicolini05,Mostaghim07}. Specifically, in the $\mathscr{S}$-metric selection evolutionary multi-objective optimization algorithm (SMS-EMOA) \citep{Beume07}, indicators such as the hypervolume measure, are therefore used as a secondary fitness after non-dominated sorting. A recently published technique to overcome this limitation of the hypervolume, without relying on domination-based properties, is to assign a fitness value to dominated solutions based on their distance to the boundary of the dominated area, i.e., the boundary surface of the hypervolume \citep{Toure19}. The resulting uncrowded hypervolume improvement (UHVI) is  not a set-based measure, and cannot be used directly to optimize approximation sets. The Sofomore framework \citep{Toure19} was therefore formulated, in which multiple optimizers are interleaved that each solve a dynamic single-objective optimization problem. In this framework,  a single approximation set is iteratively optimized, but by doing so, it loses some of the advantages of population-based evolutionary algorithms, such as the ability to escape local optima. 

In this work, we show how and when hypervolume-based MO optimization can be used to replace or supplement domination-based MOEAs. To this end, we formulate the uncrowded hypervolume (UHV) measure, which can be directly used to achieve population-based hypervolume-driven MO optimization using a single-objective problem formulation. We use the gene-pool optimal mixing evolutionary algorithm (GOMEA) \citep{bouter17,bouter17b} as baseline algorithm to compare domination-based and hypervolume based MO optimization. This algorithm has both a single-objective \citep{bouter17} and MO \citep{bouter17b} version, with recently published excellent results. In both versions, essentially, the same variation operators are used, which provides a more fair and modern comparison than when e.g., NSGA-II and the Covariance Matrix Adaptation Evolutionary Strategy (CMA-ES) \citep{hansen05} were to be used. Furthermore, in this paper we will show that the single-objective GOMEA can be used to efficiently solve the UHV problem formulation by exploiting grey-box properties that are intrinsically present in the UHV, while still assuming the MO problem itself to be black box.

The remainder of this paper is organized as follows. In Section~\ref{sec:emo20_hv}, we discuss the hypervolume and related measures. In Section~\ref{sec:emo20_gom}, we introduce an indicator-based MO optimization problem formulation based on the UHV, and how we can efficiently solve this problem with GOMEA, resulting in UHV-GOMEA. In Section~\ref{sec:emo20_sof}, we furthermore incorporate GOMEA in the Sofomore framework (Sofomore-GOMEA), and in Section~\ref{sec:emo20_mogomea}, we discuss MO-GOMEA. In Section~\ref{sec:emo20_experiments}, we empirically compare these three algorithms and discuss which of the approaches is preferred in which scenario, and how to combine them into a simple hybrid approach, on both simple benchmark problems and the commonly used WGF test suite \citep{Huband2005}. We discuss the overall results in Section~\ref{sec:emo20_discussion} and conclude in Section~\ref{sec:emo20_conclusion}.

\section{Preliminaries: The Hypervolume measure}
\label{sec:emo20_hv}
A solution $\bx\inX$ is said to \textit{weakly dominate} another solution $\by\inX$, written as $\bx\preceq\by$, if and only if $f_i(\bx) \leq f_i(\by)$ for all $i \in \{1,\ldots,m\}$. When the relation $f_i(\bx) < f_i(\by)$ is furthermore strict for at least one $i$, we say that $\bx$ \textit{dominates} $\by$, written as $\bx\prec\by$ or, with a slight abuse of notation, as $\bff(\bx) < \bff(\by)$. A solution that is not dominated by any other solution in $\cX$ is called \textit{Pareto optimal}. The \textit{Pareto set} $\cA^\star$ can then be formulated as $\cA^\star = \{ \bx\inX : \nexists \by\inX : \by\prec\bx \} \subset \cX$, while the \textit{Pareto front} is given by $\{ \bff(\bx) : \bx\in\cA^\star \} \subset \bbR^m$. 

\begin{figure}
\begin{center}
\includegraphics[width=0.7\columnwidth]{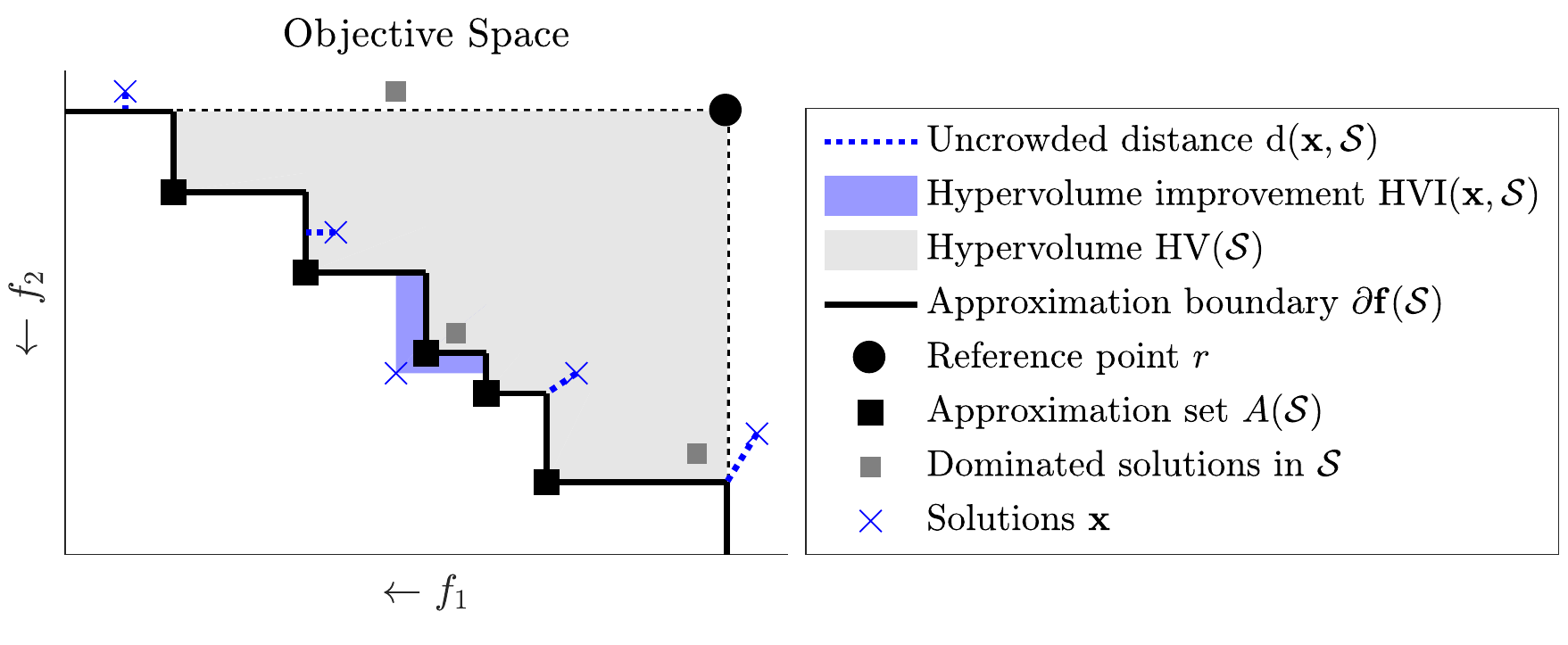}
\vspace*{-0.3cm}
\caption{Illustration of the hypervolume measure for a bi-objective minimization problem, together with the uncrowded distance and hypervolume improvement of some example solutions $\bx$ with respect to solution set $\cS$.}
\label{fig:emo20_uhv}
\end{center}
\end{figure}

Let $\powerset(\cX)$ be the powerset of $\cX$, i.e., the set of all solution sets $\cS\subseteq\cX$. The hypervolume measure $\mbox{HV} : \powerset(\cX) \rightarrow \bbR$ \citep{Zitzler99,Zitzler2003,Auger2009} of a solution set $\cS$ with respect to a reference point $r \in\bbR^m$ measures the volume dominated by all $\bx\in\cS$, and bounded by $r$, as is illustrated in Figure~\ref{fig:emo20_uhv}. Let $A : \powerset(\cX) \rightarrow \powerset(\cX)$ be the approximation set of $\cS$, $A(\cS) = \{\bx\in\cS : \bff(\bx) < r, \nexists \by\in\cS : \by\prec\bx\}$, i.e., the largest subset of $\cS$ that contains only non-dominated solutions within the region defined by $r$. Additionally, let the hypervolume improvement $\text{HVI} : \cX \times \powerset(\cX) \rightarrow \bbR$ of a solution $\bx$ with respect to a solution set $\cS$ be defined as the increase in hypervolume when $\bx$ is added to $\cS$, i.e., $\hvi(\bx, \cS) = \hypv(\cS \cup \{ \bx \}) - \hypv(\cS)$, as shown in Figure~\ref{fig:emo20_uhv}.

\begin{figure}
\begin{center}
\includegraphics[width=0.7\columnwidth]{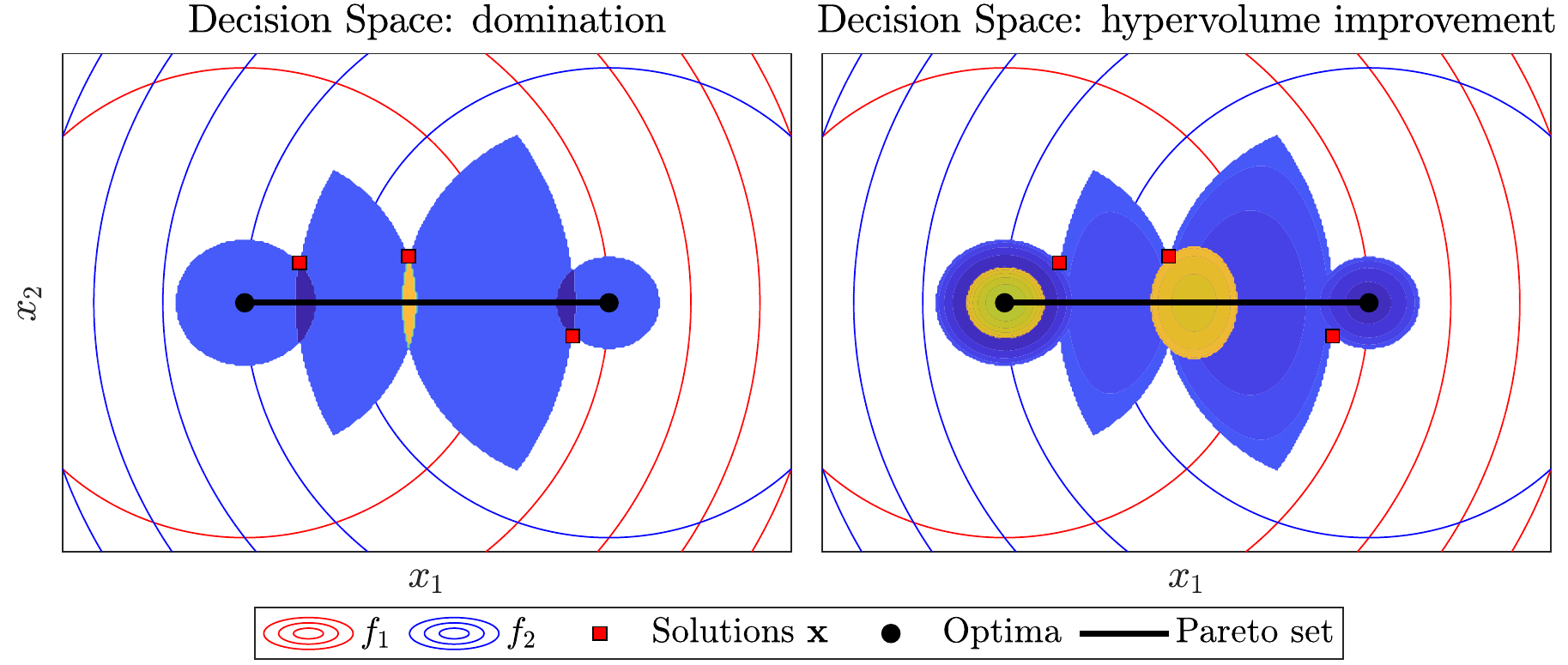}
\caption{Domination region (left) compared to the hypervolume improvement region (right) for the bi-sphere problem. Additionally, in both subfigures, the orange region is the improvement region when the middle solution is being replaced. Solutions in the light blue region are non-dominated with respect to the current solutions. The dark blue regions dominate one of the current solutions. In the right subfigure, solutions in the blue region improve the hypervolume if added to the solution set, and darker blue represents larger improvements. }
\label{fig:emo20_dom_vs_hvi}
\end{center}
\end{figure}

The difference between domination-based improvements and hypervolume-based improvements is illustrated in Figure~\ref{fig:emo20_dom_vs_hvi} for the bi-sphere problem with hypervolume reference point $r = (11,11)$. This problem is composed of two single-objective sphere problems, $f_\text{sphere}(\bx) = \sum_{i = 1}^n x_i^2,$
of which one is translated, $\bff_\text{bi-sphere}(\bx) = [ f_\text{sphere}(\bx)\; ; \;f_\text{sphere}(\bx - \be_1)],$ where $\be_i$ is the $i^\text{th}$ unit vector, with all zeros except a one in the $i^\text{th}$ position. Figure~\ref{fig:emo20_dom_vs_hvi} shows that $\text{HVI}(\bx,\cS) \geq 0 \Leftrightarrow \{\nexists \by\in\cS : \by\prec\bx\}$ holds. Indeed, both approaches have the same improvement region (in blue).
However, when comparing the \textit{improvement region for the middle solution} (in orange), i.e., when that solution is replaced in $\cS$, we see that its hypervolume improvement region is larger, essentially making it easier to find improvements. Additionally, the hypervolume improvement region consists of two disconnected subsets, indicating that it takes diversity into account, in contrast to domination-based improvements.

When $\bx$ is dominated by any solution in $\cS$, its hypervolume improvement $\text{HVI}(\bx,\cS)$ is zero. The uncrowded hypervolume improvement (UHVI) \citep{Toure19} was recently introduced to overcome this. Let $\partial \bff(\cS)$ be the \textit{approximation boundary}, i.e., the boundary between the dominated and non-dominated region in objective space, bounded by the reference point $r$, as illustrated in Figure~\ref{fig:emo20_uhv}. 
Let $\text{ud}(\bx,\cS)$ be the \textit{uncrowded distance}, which measures the shortest Euclidean distance between $\bx$ and $\partial \bff(\cS) = \partial \bff(A(\cS))$, when $\bx$ is dominated by any solution in $\cS$ or outside the region defined by $r$. 
Else, we set $\text{ud}(\bx,\cS) = 0$. It is called the uncrowded distance as the shortest distance to $\partial \bff(\cS)$ is obtained for a point on $\partial \bff(\cS)$ that is not in $\cS$ itself. The UHVI can then be defined as, $$\text{UHVI}(\bx,\cS) = \hvi(\bx,\cS)  - \text{ud}(\bx, \cS).$$

It is interesting to note that the uncrowded distance is similar to the distance measure $d^+$ that was used to construct a weak Pareto compliance version of the inverted generational distance indicator, called the $\text{IGD}^+$ \citep{Ishibuchi2015}. In the $\text{IGD}^+$, a reference set $\cZ$ of non-dominated solutions is used (i.e., a subset of the known Pareto set), and the distance of reference solutions $\bz\in\cZ$ towards the approximation boundary $\partial \bff(\cS)$ is computed. There, the solutions $\bz$ are non-dominated with respect to $\cS$. Here however, we are particularly interested in the other case, where solutions are dominated with respect to $\cS$.

\section{The Uncrowded Hypervolume}
\label{sec:emo20_gom}
The UHVI is a measure of solution quality with respect to a solution set $\cS$, but not a measure of quality for $\cS$ itself.  We therefore introduce the \textit{uncrowded hypervolume} (UHV) in this work as the hypervolume of $\cS$ penalized by all uncrowded distances,
\begin{equation}
\label{eqn:emo20_uhv}
\text{UHV}(\cS) = \text{HV}(\cS) - \frac{1}{|\cS|} \sum_{\bx\in\cS} \text{ud}(\bx, \cS)^m.
\end{equation}
The uncrowded distances are taken to the power $m$ such that they have the same unit as the hypervolume. As improving a non-dominated solution in $\cS$ could increase both its hypervolume and the uncrowded distances, the factor ${1}/|\cS|$ is added to guarantee that an improvement in hypervolume is not negatively influenced by the increase in uncrowded distances. The uncrowded hypervolume is a strictly monotonic indicator on the space of approximation sets (i.e., sets containing only non-dominated solutions), as it is equal to the hypervolume for those sets. However, on the entire space of solution sets, strict monotonicity does not hold.

\subsection{UHV-maximization}
In an indicator-based multi-objective optimization problem (IBMOP), a quality indicator is used to assign a quality value to a solution set. The underlying idea is that a single-objective optimizer then can be used to explicitly search for a solution set $\cS$ that maximizes this indicator \citep{Zitzler04,Beume07}. Let $I_\bff : \powerset(\cX) \rightarrow \bbR$ be such an indicator, with respect to the multi-objective problem given by $\bff$. To be able to search the space of solution sets, we parameterize solutions sets by considering sets of fixed size $|\cS_p| = p \geq 1$. Let $\phi$ be a vector of concatenated decision variables of the solutions in $\cS_p$, i.e., $\phi = [\bx_1 \cdots \bx_p] \in \bbR^{p\cdot n}$. Inversely, let $S(\phi) = \{\bx_1,\ldots,\bx_p\}$ be the operator that transforms $\phi$ into a solution set. To avoid confusion, we will refer to a solution $\phi$ of the IBMOP with objective function $g$ as a $g$-solution, while solutions of the original MO optimization problem given by $\bff$ are called MO-solutions from now on. IBMOPs are then formulated as,
\begin{equation}
\label{eqn:emo20_ibmop}
\begin{split}
\text{maximize} \quad & g_{\bff,p} : (\cX)^p \rightarrow \bbR, \\
\text{with} \quad & g_{\bff,p}(\phi) = I_\bff(S(\phi)), \\
& \bff : \cX \subseteq \bbR^n \rightarrow \bbR^m,  p \geq 1\\
& \phi = [\bx_1 \cdots \bx_p] \in (\cX)^p\subseteq \bbR^{p\cdot n}. \\
\end{split}
\end{equation}

This IBMOP formulation fully specifies the MO optimization problem, as well as its optimum, i.e., the resulting distribution of solutions along the front. In case of the (uncrowded) hypervolume as quality indicator, this optimal distribution known as the optimal $\mu$-distribution \citep{Auger2009a}. If and only if the front is linear, the solutions in the optimal distribution are equally spaced along the front. In general, the density of solutions is proportional to the negative slope of the front \citep{Auger2009a}.

This clear and unique definition of the optimum of the IBMOP allows us to discuss convergence to optimality, in contrast to the general aim of MO optimization that translates to a trade-off between proximity and diversity \citep{bosman03}. Furthermore, since the hypervolume measure is strictly monotonic with respect to Pareto dominance, optimality here implies that the obtained solutions are a subset of the Pareto set.

Note that the optimal distribution of solutions along the front is determined by the choice of reference point, which we fix during the course of the optimization run. If a dynamic reference point would have been used, cyclic behavior could again occur.

\subsection{UHV-GOMEA}
We use the real-valued single-objective gene-pool optimal mixing evolutionary algorithm (GOMEA) to solve IBMOPs with the UHV as indicator, which we call UHV-GOMEA. We use GOMEA as published in \citep{bouter17}, and make only minor adaptations to better align the algorithm with IBMOPs. We discuss the outline of GOMEA here, and refer the reader to \citep{bouter17} for a full description.

GOMEA is a model-based evolutionary algorithm that maintains a population of $N$ solutions. For the IBMOP these are the $g$-solutions $\phi^j\in\bbR^{p\cdot n}$, $j = 1,\ldots,N$. The variation operator in GOMEA is called  gene-pool optimal mixing (GOM). GOM was designed specifically to perform variation by adapting only a few decision variables at a time, and thereby exploiting that objective functions can often be quickly updated if only a few variables change. This notion of \textit{partial evaluations} typically requires that some problem knowledge is known, resulting in a \textit{grey-box} scenario. IBMOPs are by definition grey-box, in the sense that it is known how to update the indicator value when the decision variables corresponding to only one (or a few) MO-solution change, without having to re-evaluate all other MO-solutions, or re-compute the indicator value from scratch. This means that grey-box properties of the IBMOP can be exploited while still considering $\bff$ as a black-box.

GOMEA is equipped with a \textit{linkage model} that specifies which subsets of decision variables must be adapted simultaneously. A {linkage model} $L$ is a subset of the power set $\powerset(I)$ of all decision variables, which is $I = \{1,\ldots,pn\}$ in case of an IBMOP. A linkage model is therefore also known as a \textit{family of subsets} (FOS). In GOM, variation is performed by iteratively updating only the decision variables specified by a linkage subset $l\in L$, with $l\subseteq I$. We denote the subset of decision variables in a $g$-solution $\phi$ specified by $l$ with $\phi_{\langle l \rangle} \in \bbR^\abs{l}$. An $|l|$-dimensional Gaussian distribution is then estimated from the $\tau N$ best $g$-solutions (with $\tau = 0.35)$. From this distribution, to update each $g$-solution in the population, new values for the decision variables specified by $l$ are sampled. In GOM, only improvements to $g$-solutions are accepted. Else, the proposed update is discarded. 

Originally in GOM, the sample distributions for all linkage subsets $l\in L$ were estimated at the beginning of each generation \citep{bouter17}. Instead, here, we estimate the sample distribution for each linkage subset right before sampling new values for the variables in that subset. By doing so, the sample model is learned based on the most recent set of $g$-solutions. We found this to improve the rate of convergence, at increased computationally complexity of only $\cO((\abs{L}-1) \cdot N \log N)$ per generation, as selection now needs to be performed for each linkage subset. A Cholesky decomposition \citep{Lay93,Higham08} of the covariance matrix is then performed which is required for the sampling. This decomposition requires a positive definite covariance matrix. Due to numerical errors or small population size, it might be that the decomposition fails \citep{maree12}. If this happens once for a linkage subset, we from then on perform a regularization using the Ledoit-Wolf shrinkage estimator (LWSE) \citep{Ledoit12}. The LWSE estimates a covariance matrix based on a convex combination of the maximum likelihood estimator (MLE) and a prior matrix, for which we use the diagonal variance matrix (i.e., the diagonal of the MLE). Additionally, when $|l| > \tau N - 1$ , we always use a diagonal variance matrix without attempting a decomposition.

\subsubsection{IBMOP linkage models}
\label{sec:emo20_linkage_models}
Any linkage model can be used in GOMEA, but we discuss three models that we employ in UHV-GOMEA in this work. Let $I_{(i)}= \{ \phi_{(i-1)n + 1},\ldots,\phi_{in} \}$ for $i = 1,\ldots,p$ be the decision variables corresponding to the MO-solution $\bx_i$ of $g$-solution $\phi = [\bx_1\cdots\bx_p]$. The three linkage models we consider are then given by,
\begin{equation}
\label{eqn:emo2_L}
\begin{split}
Lm =		&\, \left\{ I_{(1)}, I_{(2)}, \ldots, I_{(p)} \right\}, \\
Lf  = 	&\, \left\{ I_{(1)} \cup I_{(2)} \cup \cdots \cup I_{(p)} \right\} = I, \\
Lt  =	&\,  \text{UPGMA}(Lm). \\
\end{split}
\end{equation}

In the \textit{marginal linkage model} $Lm$, only decisions variables corresponding to the same MO-solution are considered to be dependent, i.e., $\phi_{\langle l_i \rangle} = \bx_i$ for $l_i\in Lm$. In the \textit{full linkage model} $Lf$, the linkage model contains only a single subset, in which all decision variables are considered to be dependent, i.e., $\phi_{\langle l \rangle} = \phi_{\langle I \rangle} = \phi$. Finally, the \textit{linkage tree model} $Lt$ contains multiple levels of linkage, constructed by hierarchically clustering the subsets of the marginal model $Lm$ using UPGMA \citep{Gronau07}, where the two \textit{nearest} subsets are merged iteratively and added to $Lt$. Since each subset in $l_i \in Lm$ corresponds to the parameters of a single MO-solution $\bx_i$, we can compute the mean objective values of each $l_i$, $\bm_i = \frac1N\sum_{j = 1}^N \bff\left(\phi_{\langle l_i \rangle}^j\right) = \frac1N\sum_{j=1}^N\bff\left(\bx_i^j\right)$. We use the distance between these means as distance measure between linkage subsets. Since these change over time, the linkage tree is re-constructed every generation. Linkage tree construction results in $|Lt| = 2p-1$ subsets, and $Lt$ always contains all elements of $Lm$ as well as $Lf$, but the other subsets depend on the merge order defined by this distance measure. Additionally, in GOM, all linkage subsets for which $|l| > \tau N - 1$ are skipped, as for these relatively large linkage subsets, dependencies cannot be estimated, as a full-rank covariance matrix cannot be estimated.

\subsubsection{Permuting MO-solutions}
Since permuting solutions in $\cS$ does not influence its corresponding hypervolume, we aim to reorder the MO-solutions in $S(\phi)$, and therefore $g$-solution $\phi$, at the beginning of a generation, such that all $\phi_{\langle l_i\rangle}^j = \bx_i^j$ for each $g$-solution $j = 1,\ldots,N$ belong to a similar part of the approximation front. Therefore, we again compute the objective-space means $\bm_i$ for $i = 1,\ldots,p$. For each $g$-solution $\phi^j$, the MO-solutions are re-ordered in a greedy fashion by iteratively finding the (next-)nearest MO-solution-mean pair. 

\subsubsection{Elitist archive}
\label{sec:emo20_archive}
An elitist archive $\cE \subset \cX $ is maintained that contains all non-dominated MO-solutions $\bx\inX$ that were evaluated during optimization. To keep the archive size tractable, adaptive objective space discretization was used as presented in \cite{luong12}. If, with the archiving scheme, the archive exceeds the target size of $N_\cE = 1000$ MO-solutions, the objective space is discretized into boxes, and only one MO-solution per box is maintained. Newly obtained non-dominated MO-solutions are then only added to the archive when they end up in an empty objective-space box, or when they dominate the MO-solution in that box. This archiving scheme is also used in MO-GOMEA \citep{bouter17b}, and the target size of the elitist archive $\cE$ is set to $N_\cE = 1000$ MO-solutions. The elitist archive is not an essential part of UHV-GOMEA, but merely added to allow for a comparison to archive-based MOEAs.

\subsection{Generational Computational Complexity}
A main limitation of the usage of the hypervolume measure in the optimization process is its computational complexity. Computation of the UHV of a solution set $\cS_p$ containing $p$ MO-solutions consists of three steps. First, the set of non-dominated solutions $A(\cS_p)$ needs to be constructed, which can be performed in $\cO(mp^2)$ time. Then, the hypervolume of $A(\cS_p)$ needs to be computed. In the case of $m = 2$ two objectives, this can be done by simply sorting the solutions, resulting in a computational complexity of $\cO(p\log p)$ time. Finally, using the sorted approximation set, computing the uncrowded distances can be performed in $\cO(mp)$ time. This results in an overall computational complexity of $\cO(mp^2)$ for computing the UHV with $m = 2$ objectives. Note however that $p$ is typically rather small here, (e.g., $p  = 9$).  Note further that we did not yet explore potential speedups by considering that the hypervolume measure does not have to be computed from scratch every generation, but can be updated.

In contrast, ``Fast Non-Dominated Sorting" in NSGA-II with a population of $N$ MO-solutions can be performed in $\cO(mN^2)$ time \citep{deb02}, however, there, $N$ is typically larger (e.g., $N = 100$). Also, maintaining an elitist archive of size $N_\cE \leq 1000$ has a computation complexity of $\cO(mN_\cE)$, and is essential for the performance of MO-GOMEA \citep{bouter17b}.

To compute the hypervolume with $m \geq 3$ objectives, different algorithms have been proposed \citep{Fonseca2006,Beume2009}, yielding a computational complexity of $\cO(p^{m-2}\log p)$, which becomes the dominating term when $m$ is large. Additionally, an efficient algorithm to compute the uncrowded distance for $m > 2$ has still to be derived.

\section{Sofomore-GOMEA}
\label{sec:emo20_sof}
While UHV-GOMEA optimizes the UHV by manipulating sets of MO-solutions in a population-based approach, the Sofomore framework \citep{Toure19} was introduced to iteratively optimize the UHVI one MO-solution at a time. Specifically, Sofomore performs a search around a single solution set $\cS_p$ of fixed size $p$. For each MO-solution $\bx_i \in\cS_p$, a single-objective optimizer is initialized that solves the single-objective dynamic optimization problem $h_i : \cX \rightarrow \bbR$ given by  $h_i(\bx_i|\cS_p\backslash\{\bx_i\}) = \text{UHVI}(\bx_i,\cS_p\backslash\{\bx_i\})$. Note that the decision space of $h_i$ and the MO problem given by $\bff$ are the same, and an $h_i$-solution is thus also an MO-solution, but with different fitness value assignment. Steps of individual optimizers are then interleaved. Intuitively, each optimizer in turn aims to replace $\bx_i$ by the solution with maximal hypervolume contribution with respect to $\cS_p$, while keeping the other MO-solutions in $\cS_p$ fixed.

To solve the $p$ dynamic single-objective optimizations problems given by $h_i$, we again use GOMEA to get a fair comparison basis for comparing different optimization approaches. Our implementation largely agrees with the combination of Sofomore and CMA-ES (called COMO-CMA-ES), as presented in \citep{Toure19}, but a few minor changes were made that we discuss here. We refer the reader to \citep{Toure19} for further details on Sofomore and COMO-CMA-ES.

We assumed that the MO problem is black-box, and we therefore use a full linkage model for Sofomore-GOMEA, as $h_i$ cannot be partially evaluated. We maintain the default uniform initialization of GOMEA. In this way, Sofomore-GOMEA and HV-GOMEA are initialized from exactly the same set of MO-solutions (when the same random seed is used). Additionally, and most important, the Sofomore approach results in a set of \textit{dynamic} optimization problems. As GOMEA accepts only improvements to solutions, it is beneficial to recompute all fitness values (i.e., $h_i$) at the beginning of each generation, as they might have changed since the last generation. Note that this requires $h_i(\bx|\cS_p\backslash\{\bx_i\}) = \text{UHVI}(\bx_i,\cS_p\backslash\{\bx_i\})$ to be updated, but there is no need to re-compute $\bff(\bx_i)$ itself. In this work, we evaluate performance based on the number of MO function evaluations (MO-fevals), which is thus unaffected by this re-evaluation, although overall computation time will increase. This re-evaluation is not required for UHV-GOMEA, as it does not solve a dynamic problem. 
Finally, we use the same archive for Sofomore-GOMEA as described in Section~\ref{sec:emo20_archive}. 

\section{MO-GOMEA}
\label{sec:emo20_mogomea}
MO-GOMEA is a domination-based MOEA  \citep{bouter17b}. MO-GOMEA optimizes a population of MO-solutions that is aimed to approximate the Pareto front by balancing diversity and proximity. Besides the main population, MO-GOMEA maintains an elitist archive $\cE$, as described in Section~\ref{sec:emo20_archive}. In MO-GOMEA, MO-solutions are copied back from the archive into the population. It can therefore be roughly said that the aim of MO-GOMEA is to obtain an elitist archive that approximates the Pareto front as good as possible. We discuss the main characteristics of MO-GOMEA here, and refer the reader to \citep{bouter17b} for a full description of the algorithm. 

Again, as we assumed the MO problem to be black-box, we use a full linkage model for MO-GOMEA. From a population of $N_{mo}$ MO-solutions, truncation selection is performed based on domination rank, resulting in a selection of size $\tau N$. This selection is clustered into  $K_{mo}$ clusters, each of size $2\tau N / K_{mo}$. Each cluster models a part of the approximation front, and for this an objective-space based clustering method is used that guarantees overlapping clusters of equal size. For each cluster, a Gaussian distribution is estimated to sample new MO-solutions from. Similar to the single-objective GOMEA, MO-GOMEA only accepts improvements, meaning that offspring need to either dominate the parent, or be accepted into the elitist archive. 

To align MO-GOMEA with the other algorithms, we set $N_{mo} = p\cdot N$ and $K_{mo} = 2p$ such that the overall number of MO-solutions in the populations is the same, and all sample distributions are estimated from the same number of MO-solutions. MO-GOMEA estimates its sample distributions in a similar fashion as the single-objective GOMEA which was used in UHV-GOMEA and Sofomore-GOMEA. This makes a comparison between these three approaches most fair. Finally, to be able to compare the limited-size $|\cS_p| = p$ of UHV-GOMEA and Sofomore-GOMEA, we perform greedy hypervolume subset selection (gHSS) \citep{Guerreiro16} to select $p$ solutions from the elitist archive $\cE$.

\subsection{A Hybrid method}
MO-GOMEA is expected to perform better initially, but to stagnate in terms of proximity to the Pareto set when the majority of MO-solutions in the population is non-dominated. We construct a simple hybrid approach where we initially run MO-GOMEA, which we terminate when it stagnates, i.e., when $90\%$ of the MO-solutions in the population are non-dominated, or when the elitist archive target size is hit. We then switch to UHV-GOMEA-Lm starting from the elitist archive $\cE$ that MO-GOMEA obtained so far. $\cE$ is clustered into $p$ clusters of equal size $2|\cE|/p$ using the same clustering method that is used in MO-GOMEA. For this, the cluster means are initialized with gHSS, and distances are measured in decision space. If the cluster size is smaller than the desired population size, i.e., $2|\cE|/p < N$, the remainder of the MO-solutions is sampled uniformly.

\section{Experiments}
\label{sec:emo20_experiments}
\subsection{Experiment 1: Rate of Convergence}
In Section~\ref{sec:emo20_hv}, we introduced the bi-sphere problem, which we use to demonstrate the rate of convergence for UHV-GOMEA with the three linkage models, Sofomore-GOMEA, and MO-GOMEA. The bi-sphere problem is a separable problem, defined on the entire real space $\cX = \bbR^n$. Its Pareto set is a straight line between the origin and $\be_1$. Due to its separability, it can be solved with a diagonal variance matrix, and we therefore use a small population size $N = 31$. All algorithms are initialized from the same initial population (for the same random seed) on the domain $[-100,-50]^n$, away from the Pareto set.  We use $n = 10$ MO-decision variables with $p = 9$ MO-solutions in a solution set, resulting in an IBMOP with 90 decision variables.

As performance indicator, we use the distance to the optimal hypervolume $\Delta\text{HV}_p = \text{HV}(\cA^\star_p) - \text{HV}(\cA_p)$, with $\cA_p = A(\cS_p)$, and $\cA_p^\star$ the approximation set of size $p$ with optimal hypervolume \citep{Auger2009}. We determined $\mbox{HV}(\cA^\star_p)$ empirically by solving lower-dimensional problem instances with a large computational budget.  Additionally, we use the generational distance (GD) \citep{Zitzler2003}, $$\text{GD}(\cA_p,\cA^\star) = \frac{1}{|\cA_p|} \sum_{\bx\in\cA_p} \min_{\by\in\cA^\star} \norm{\bff(\bx) - \bff(\by)}.$$ Since a parametric expression of the Pareto set $\cA^\star$ is available for the bi-sphere problem, we can compute the GD analytically. The GD measures the proximity of $\cA_p$ to the Pareto set, but does not take diversity into account. The GD is not Pareto compliant, but since we considered fixed-size approximation sets, it is a useful tool to measure proximity to the Pareto set. 
We are especially interested in the GD at the end of the optimization process, when all approximation sets are the same size and contain only non-dominated solutions. Since we compute the GD analytically, it does hold that any solution set containing only Pareto optimal solutions has $\text{GD} = 0 $. Because of this property, the analytic GD is a good measure for proximity to the Pareto set. Finally, we count the non-dominated MO-solutions in $\cS_p$, i.e., $|\cA_p|$, and visualize the approximation fronts. In all experiments, we measure performance in terms of the number of MO-fevals. All experiments in this work are repeated 30 times, and mean results are shown, unless mentioned otherwise. The hypervolume reference point is set to $r = (11,11)$ in all experiments, which is far away from the Pareto front, thereby aiming that the endpoints of the Pareto front are in $\cA^\star_p$. However, even by setting the reference point this far, this is not always achieved for all problems.

\subsubsection{Results}
\begin{figure}
\begin{center}
\includegraphics[width=0.8\columnwidth]{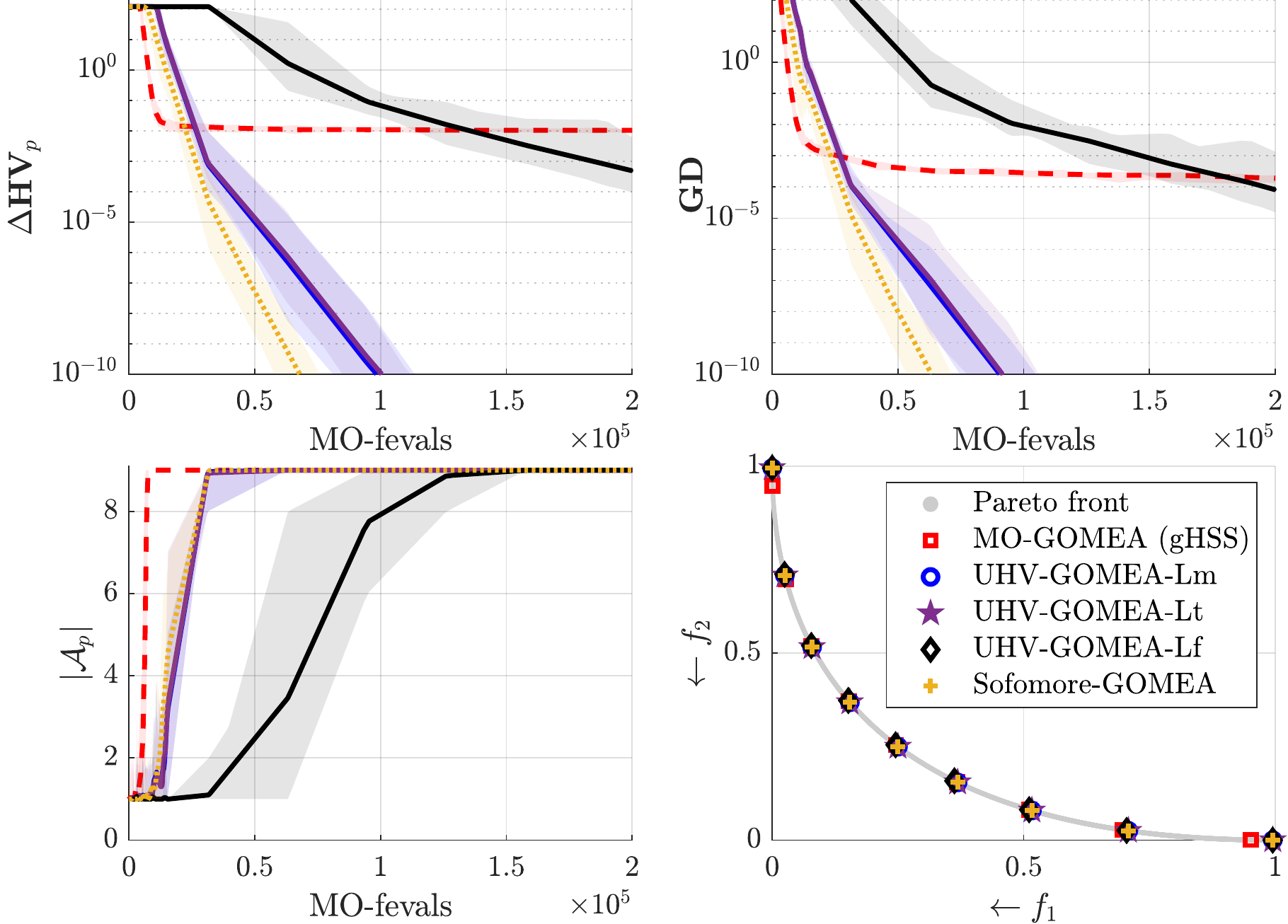}
\caption{Convergence on the bi-sphere problem (with $n = 10$, $p = 9$, and $N = 31$). Lines show mean scores over 30 independent runs, and shaded areas show min/max scores. The objective space plot (bottom right) shows the result of a single run of each algorithm. The GD is computed analytically.}
\label{fig:emo20_bisphere_experiment1}
\end{center}
\end{figure}

Figure~\ref{fig:emo20_bisphere_experiment1} shows that all hypervolume-based algorithms exhibit linear convergence in terms of $\log(\Delta\text{HV}_p)$ and $\log(\text{GD})$, albeit at different rates. Sofomore-GOMEA performs best, closely followed by UHV-GOMEA-Lm. Due to the small population size, large linkage elements are filtered from the linkage tree in UHV-GOMEA-Lt, which performs the same as UHV-GOMEA-Lm (up to randomness). UHV-GOMEA-Lf can still solve the problem, but it is inefficient. With full linkage, all MO-solutions are updated and evaluated simultaneously (albeit independently from the diagonal variance matrix, due to the small population size), and only then the corresponding hypervolume is computed. With marginal linkage however, MO-solutions are updated and evaluated one-by-one, and after each newly evaluated MO-solution, it is checked if this improved the corresponding hypervolume, which is beneficial here.

In this scenario, where the optimization is initialized far from the Pareto set, the hypervolume-based algorithms are initially fully driven by the uncrowded distance towards the reference point. The uncrowded distance within UHV-GOMEA and Sofomore-GOMEA is effective in obtaining a set of non-dominated solutions, but is not as efficient as MO-GOMEA, which performs best initially in all three measures. The rate of convergence for the hypervolume-based algorithms is constant as soon as $|\cA_9| = 9$ for UHV-GOMEA and Sofomore-GOMEA, which shows that linear convergence is due to the hypervolume optimization and not due to the uncrowded distance.

As expected, MO-GOMEA stagnates in terms of $\Delta\text{HV}_p$, which happens at $\Delta\text{HV}_p \approx 0.01$. Its elitist archive contains a good distribution of solutions along the front, including solutions close to the endpoints of the Pareto set, but its approximation front has a slightly different distribution of solutions due to the gHSS, which could explain the stagnation in $\Delta\text{HV}_p$. However, the analytic GD plot shows that these solutions do not converge to the Pareto set. 

\subsection{Experiment 2: Modeling dependencies}
The bi-sphere problem could be solved efficiently with a small population size $N$ and without dependency modeling. We now perform a number of experiments to investigate if there are scenarios where modeling the dependencies between MO-solutions is beneficial. For this, we use two additional benchmark problems, constructed from two well-known single-objective functions, $f_\text{elli}(\bx)  = \sum_{i = 1}^n 10^{6\frac{i-1}{n-1}}x_i,$ and $f_\text{Rosenbrock}(\bx)  = \sum_{i = 1}^{n-1} (  100(x_{i+1} - x_{i}^2)^2 + (1-x_i)^2)$. These are, similar to the bi-sphere problem, defined for a scalable number of decision variables $n$ and on $\cX = \bbR^n$. From these, we construct two bi-objective optimization problems,
\begin{equation}
\label{eqn:emo20_mo_functions}
\begin{split}
\bff_\text{sphere-rotatedElli}(\bx) & = \left[ f_\text{sphere}(\bx) \; ; \; f_\text{elli}(R \bx - \be_1)\right], \\
\bff_\text{sphere-Rosenbrock}(\bx) & = \left[ \frac{1}{n}f_\text{sphere}(\bx) \; ; \; \frac{1}{n-1} f_\text{Rosenbrock}(\bx)\right], \\
\end{split}
\end{equation}
where $R$ is a rotation matrix that defines a rotation around the origin of $\pi/4$ radians in all principal directions. The sphere-rotatedElli problem has the same Pareto front as the bi-sphere problem (but a different, non-linear, Pareto set), and has one non-separable ill-conditioned objective. The sphere-Rosenbrock problem (also known as BD2s \citep{bosman13}) has pair-wise dependencies and a non-linear Pareto set. Especially the combination of an easy and a difficult objective can make it difficult to obtain an evenly spread approximation front. All three problems are scaled such that their Pareto fronts have endpoints at $(1,0)$ and $(0,1)$. The sphere-rotatedElli problem is initialized on the domain $[-100,-50]^n$, the sphere-Rosenbrock function on $[-5,5]^n$. 

\subsubsection{Results: Linkage tree versus random linkage}
We compared the linkage tree model, where linkage subsets are formed between neighboring solutions (see Section~\ref{sec:emo20_linkage_models}), with a model in which solutions are merged randomly. Results were compared for 30 runs on sphere-Rosenbrock, with a population size of $N = 200$, $n = 10$ and $p = 9$. The mean number of function evaluations that UHV-GOMEA-Lt required to obtain $\Delta\mbox{HV}_p < 10^{-10}$ was $1.28\times10^6$ (range: $1.02$--$1.66\times 10^6$). When we use the same UPGMA method of constructing the linkage tree, but with a random distance between linkage models, the required number of function evaluation increases significantly to $1.59\times 10^6$ (range: $1.35$--$1.95\times 10^6$), as tested with a Wilcoxon rank-sum test with $\alpha = 0.05$. This shows that the intuitively chosen construction approach for the linkage tree has added value, although better linkage models might of course still exist.

\subsubsection{Results: Population size $N$}
\label{sec:emo20_experiment2_results}
\begin{figure}[t]
\includegraphics[width=\columnwidth]{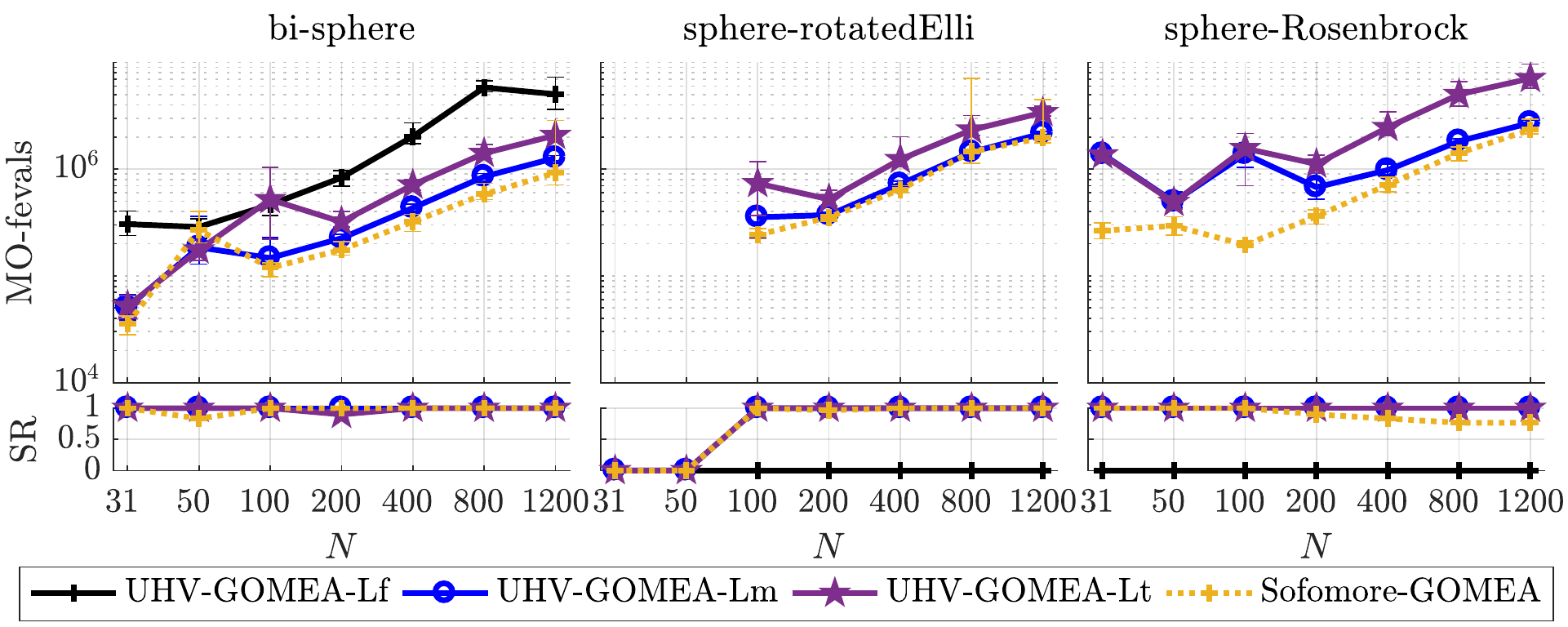}
\caption{Effect of the population size parameter $N$ on the hypervolume-based algorithms for various problems with $n=10$ and $p = 9$. Success rate (SR) measures the fraction of runs in which the target accuracy ($\Delta\mbox{HV}_p < 10^{-5}$) was reached, out of 30 runs in total. The top row shows the number of MO-fevals required to reach this accuracy, for all successful runs. }
\label{fig:emo20_experiment2_popsize}
\end{figure}

The effect of the population size parameter $N$ on the required number of MO-fevals to reach a target accuracy of $\Delta\mbox{HV}_p < 10^{-5}$, and the corresponding success rate (SR) is shown in Figure~\ref{fig:emo20_experiment2_popsize}. The computational budget was set to $10^7$ MO-fevals. Performance is rather predictable for $N \geq 200$, but for small population sizes, differences occur. UHV-GOMEA with Lt and Lm perform similar initially, as larger linkage subsets are filtered out. When increasing the population size, UHV-GOMEA-Lt performs slightly worse, although the overhead seems to be a constant factor. This suggests that it is nor beneficial nor harmful to model dependencies in this scenario. UHV-GOMEA-Lf is not able to solve the sphere-rotatedElli, and is significantly slower for the bi-sphere problem. We therefore omit it from further experiments. 

Sofomore-GOMEA performs similar to UHV-GOMEA-Lm for larger population sizes, but for smaller population sizes, it clearly outperforms all UHV-GOMEA variants on sphere-Rosenbrock.  A relatively large population size is required to estimate a full covariance matrix, which is required to solve sphere-rotatedElli. For smaller population sizes, a diagonal variance matrix or regularized covariance matrix is estimated, which affects performance, making it hard to predict whether it is beneficial to increase or decrease the population size. Interestingly, for large $N$, the success rate of Sofomore-GOMEA deteriorates on sphere-Rosenbrock, as it converges to a locally optimal distribution of solutions along the front, which can only be escaped if multiple solutions were to be updated simultaneously. This only happens for sphere-Rosenbrock due to the shape of its Pareto front, but also since the population is initialized close to the Pareto set.  Since $f_1$ is easier to solve than $f_2$,  solutions initially quickly converge towards one end the front, and the larger the population size, the faster this happens, resulting in little to no opportunity for each of the optimizers to adapt for the dynamic nature of its objective function.



\subsubsection{Results: Solution set size $p$}
\begin{table}
\smaller
\caption{Success rate (SR) to obtain $\Delta\mbox{HV}_p < 10^{-10}$, together with the MO-fevals per $p$ of successful runs ($\pm$ standard deviation) for the sphere-rotatedElli benchmark problem ($n = 3$), for $N = 50$ and $N=100$. Bold scores are best obtained scores or those not statistically different from it.}
\vspace{1em}
\label{tab:emo20_p}
\begin{center}
\begin{tabular}{l|c|lr|lr|lr}
\toprule
& & \multicolumn{2}{c|}{UHV-GOMEA-Lm} &\multicolumn{2}{c|}{UHV-GOMEA-Lt} & \multicolumn{2}{c}{Sofomore-GOMEA} \\ 
&  $p$ & SR & MO-fevals/$p$ & SR & MO-fevals/$p$ & SR & MO-fevals/$p$ \\
  \midrule
  \multirow{6}{*}{\rotatebox[origin=c]{90}{$N = 50$}}
&   3  & \textbf{\textit{1.00}} & {3.3e+03}$\pm$1.8e+02  & \textbf{\textit{1.00}} & {5.2e+03}$\pm$7.2e+02  & \textbf{\textit{1.00}} & \textbf{2.9e+03}$\pm$3.2e+02 \\ 
&   5  & \textbf{\textit{1.00}} & {4.3e+03}$\pm$2.1e+02  & \textbf{\textit{1.00}} & {9.1e+03}$\pm$1.7e+03  & \textbf{\textit{1.00}} & \textbf{3.6e+03}$\pm$4.1e+02 \\ 
&   9  & \textbf{\textit{1.00}} & \textbf{7.2e+03}$\pm$5.8e+02  & \textbf{\textit{1.00}} & {1.6e+04}$\pm$2.1e+03  & \textbf{\textit{1.00}} & {7.5e+03}$\pm$3.0e+02 \\ 
&  17  & \textit{0.73} & \textbf{2.4e+04}$\pm$7.5e+03  & \textit{0.90} & {8.2e+04}$\pm$1.9e+04  & \textbf{\textit{1.00}} & \textbf{2.2e+04}$\pm$7.0e+02 \\ 
&  33  & \textit{0.00} & -  & \textbf{\textit{0.30}} & \textbf{3.2e+05}$\pm$1.6e+05  & \textit{0.00} & - \\ 
&  65  & \textit{0.00} & -  & \textit{0.00} & - & \textit{0.00} & - \\

   \midrule
\multirow{6}{*}{\rotatebox[origin=c]{90}{$N = 100$}}     
&   3  & \textbf{\textit{1.00}} & {6.5e+03}$\pm$2.4e+02  & \textbf{\textit{1.00}} & {8.5e+03}$\pm$5.9e+02  & \textbf{\textit{1.00}} & \textbf{5.3e+03}$\pm$4.8e+02 \\ 
&   5  & \textbf{\textit{1.00}} & {8.3e+03}$\pm$2.2e+02  & \textbf{\textit{1.00}} & {1.3e+04}$\pm$9.9e+02  & \textbf{\textit{1.00}} & \textbf{6.1e+03}$\pm$4.2e+02 \\ 
&   9  & \textbf{\textit{1.00}} & \textbf{1.3e+04}$\pm$6.4e+02  & \textbf{\textit{1.00}} & {2.4e+04}$\pm$2.8e+03  & \textbf{\textit{1.00}} & {1.4e+04}$\pm$8.5e+02 \\ 
&  17  & \textit{0.63} & \textbf{3.5e+04}$\pm$7.3e+03  & \textbf{\textit{1.00}} & {4.0e+04}$\pm$3.5e+03  & \textbf{\textit{1.00}} & {4.1e+04}$\pm$1.7e+02 \\ 
&  33  & \textit{0.00} & - & \textbf{\textit{0.97}} & \textbf{1.5e+05}$\pm$3.9e+04  & \textit{0.23} & \textbf{1.4e+05}$\pm$4.0e+03 \\ 
&  65  & \textit{0.00} & -  & \textbf{\textit{0.97}} & \textbf{8.3e+05}$\pm$2.4e+05  & \textit{0.00} & - \\ 

 \bottomrule
\end{tabular}
\end{center}
\end{table}
When the size $p$ of the solution set $\cS_p$ is set larger, resulting MO-solutions will be closer to each other on the front. It can thus be expected that for larger $p$, dependency modeling becomes more relevant. We let $p = 2^j + 1$ with $j \in \bbN$, and inspect how many runs obtain a high accuracy of $\Delta\mbox{HV}_p < 10^{-10}$. For this, we let each algorithm run with a large budget of $10^8$ MO-fevals, or until it converged, i.e., when the standard deviation of the objective values in the population is less than $10^{-20}$. At this point, the machine accuracy becomes an issue, and no further improvements can be obtained. We use the sphere-rotatedElli problem with $n = 3$, as it has dependencies, with the same Pareto front as the bi-sphere problem, so that we can find the empirical $\Delta\mbox{HV}(\cA_p^\star)$ by solving this simpler problem with accuracy of $10^{-15}$. Success rates and number of MO-fevals to reach the target accuracy are shown in Table~\ref{tab:emo20_p}. Differences were tested for statistical significance using the Wilcoxon rank-sum test at $\alpha = 0.05$.

For $N=50$ and $p\leq 9$, all algorithms obtain the target accuracy in all runs, but for $p\geq17$, performance deteriorates for UHV-GOMEA-Lm and Sofomore-GOMEA, while UHV-GOMEA-Lt can still solve the problem. Doubling the population size roughly doubles the required number of MO-fevals for all algorithms for $p\leq 9$. For UHV-GOMEA-Lt, increasing the population size allows it to solve problems with large $p$, clearly outperforming the other algorithms. 

The number of function evaluations GOMEA requires to solve grey-box problems with independent decision variables scales logarithmically in the problem dimensionality \citep{bouter17}, which is for the IBMOP determined by $p$ and $n$. Here, we see that the required number of MO-fevals grows super-linear in $p$. This indicates that the strength of the dependencies indeed increases with $p$, up to the point that UHV-GOMEA-Lm and Sofomore-GOMEA can no longer solve the problem, making dependency modeling essential.

\subsection{Experiment 3: Elitist archive}
In Experiment 1, the limited-size $\cS_p$ obtained by the different algorithms was the basis of comparison together with a notion of proximity of each MO-solution to the Pareto set, which was favorable for the hypervolume-based algorithms. We now compare the elitist archives $\cE$. All algorithms have the same archiving strategy, but only MO-GOMEA uses the archive as part of the optimization process. As performance indicator, we use the inverse generational distance (IGD) \citep{bosman02}, $$\text{IGD}(\cE,\cA^\star) = \frac{1}{|\cA^\star|} \sum_{\by\in\cA^\star} \min_{\bx\in\cA} \norm{\bff(\bx) - \bff(\by)},$$ 
where we use a finite approximation of the Pareto set $\cA^\star \approx \cA_{5000}^\star$ of 5000 MO-solutions. The IGD measures both proximity and diversity, in contrast to the GD, which only measures proximity. We run each algorithm until a target accuracy of IGD $<10^{-3}$ is obtained, or when it converged before.

\subsubsection{Results}
\begin{figure}[t]
\begin{center}
\includegraphics[width=0.9\columnwidth]{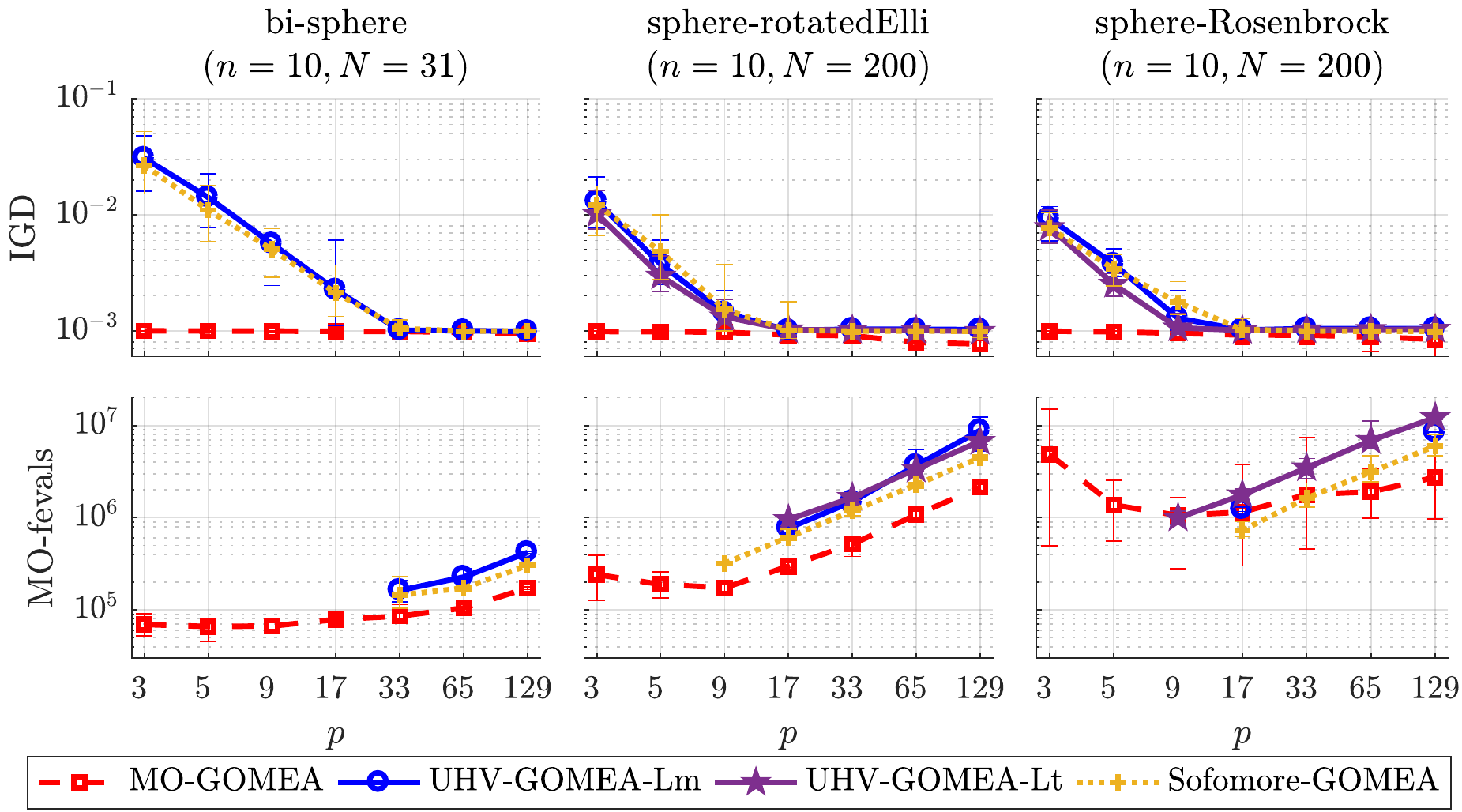}
\caption{Inverse generational distance (IGD) and number of MO-fevals to reach $\text{IGD}<10^{-3}$ in case of success.}
\label{fig:emo20_experiment4_archive}
\end{center}
\end{figure}

Results are shown in Figure~\ref{fig:emo20_experiment4_archive}. Independent of the choice of $p$, MO-GOMEA was able to obtain the target IGD. Performance seems rather robust against the choice of $p$, which, in our experiment setup controls its population size $N_{mo} = p\cdot N$ and number of clusters $K_{mo} = 2p$. Furthermore, we see that MO-GOMEA outperforms the other algorithms in all cases by obtaining a better IGD with fewer MO-fevals, except for some runs of the sphere-Rosenbrock problem with $p = 17$ and $p = 33$. In terms of $p$, $\log(\text{IGD})$ decreases linearly for the hypervolume-based algorithms, and for $p\geq 33$, all algorithms obtain the target IGD, for all three problems.

\subsection{Experiment 4: Multimodal MO problems}
As the Sofomore framework performs a form of local search around a single approximation set, it is as expected that it performs well on unimodal functions. Real-world problems are often not that well-behaved. Therefore, we now include two multimodal problems from the ZDT problem set \citep{Deb02test}. ZDT3 is characterized by a discontinuous Pareto front. ZDT6 has a concave front, and six local optima in $f_1$, resulting in a Pareto set consisting of six subsets that partially overlap. The decision space of the ZDT problems is bounded to $[0,1]^n$. We here use re-sampling as repair mechanism. We furthermore set $p = 9$ and therefore use $N = 200$, as this was found to be a good choice for all algorithms based on the results in Section~\ref{sec:emo20_experiment2_results}.

\subsubsection{Results} 
\begin{figure}[t]
\begin{center}
\includegraphics[width=0.9\columnwidth]{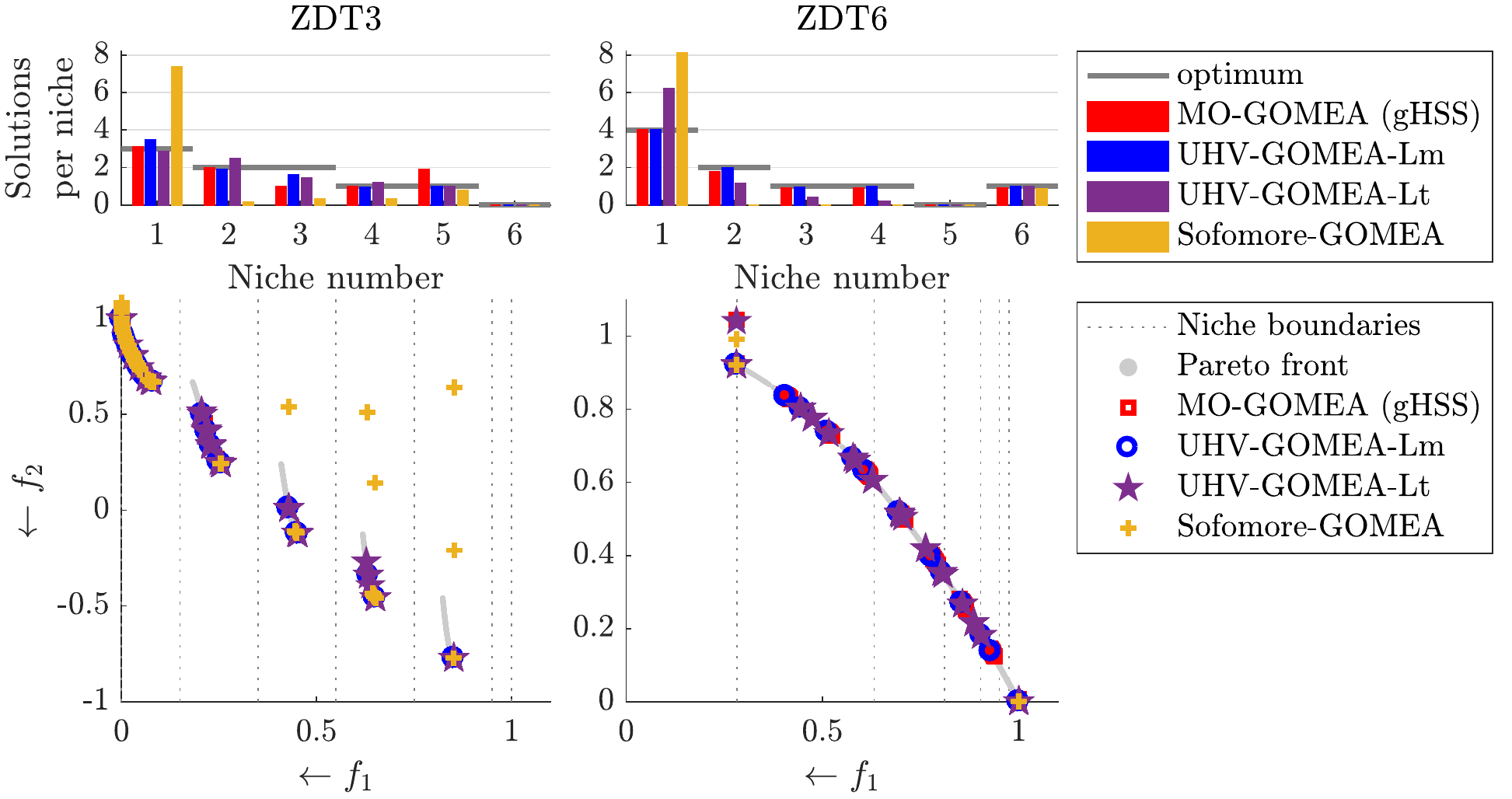}
\caption{Distribution of solutions over different niches in the final solution sets $\cS_p$ for the multimodal functions ZDT3 and ZDT6 ($n = 10$, $p = 9$, $N = 200$). Niche numbers correspond to the objective space regions separated by the dotted lines.}
\label{fig:emo20_experiment5_mm}
\end{center}
\end{figure}

Figure~\ref{fig:emo20_experiment5_mm} shows the distribution of MO-solutions per niche. UHV-GOMEA-Lm obtains the optimal hypervolume for all runs on ZDT6 and for 60\% of the runs on ZDT3. MO-GOMEA obtains an elitist archive with MO-solutions in all niches, but gHSS does not result in the optimal distribution over niches. Sofomore-GOMEA obtains many more MO-solutions in the left-most niche (Niche 1). It quickly converges to the extremes of the Pareto front in the beginning of the optimization, but is not able to move MO-solutions along the front. For ZDT3, a good spread of MO-solutions within the left-most niche is obtained, but it is not able to move MO-solutions out of that local optimum to the other subsets of the Pareto front in most runs. When it did not have MO-solutions in one of the subsets of the front, it sometimes ended up with non-optimal MO-solutions within the niches that were obtained in an attempt to fill the resulting gaps in the front. For ZDT6, Sofomore-GOMEA shows the same behavior, but it is not able to move MO-solutions along the front due to the concavity of the front, and only obtains MO-solutions at the extremes of the Pareto front. UHV-GOMEA-Lm obtains $\Delta\text{HV}_p < 10^{-5}$ in 60\% of the runs for ZDT3 and 100\% for ZDT6, while UHV-GOMEA-Lt obtains this accuracy 26\% of the runs for ZDT3, and 0\% for ZDT6. The other methods were not able to reach this accuracy in any of the runs.

\subsection{Experiment 5: A hybrid approach}
As seen in Figure~\ref{fig:emo20_bisphere_experiment1}, MO-GOMEA performs best initially, but stagnates in terms of proximity (e.g., GD) when the majority of MO-solutions in the population is non-dominated. We construct a simple hybrid approach where we initially run MO-GOMEA, which we terminate when it is expected to stagnate, i.e., when $90\%$ of the MO-solutions in the population are non-dominated, or when the elitist archive target size is hit. We then switch to UHV-GOMEA-Lm starting from the elitist archive $\cE$ that MO-GOMEA obtained so far. $\cE$ is clustered into $p$ clusters of equal size $2|\cE|/p$ using the same clustering method that is used in MO-GOMEA. For this, the cluster means are initialized with gHSS, and distances are measured in decision space. If the cluster size is smaller than the desired population size, i.e., $2|\cE|/p < N$, the remainder of the MO-solutions is sampled uniformly.

\subsubsection{Results}
\begin{figure*}[t]
\includegraphics[width=\textwidth]{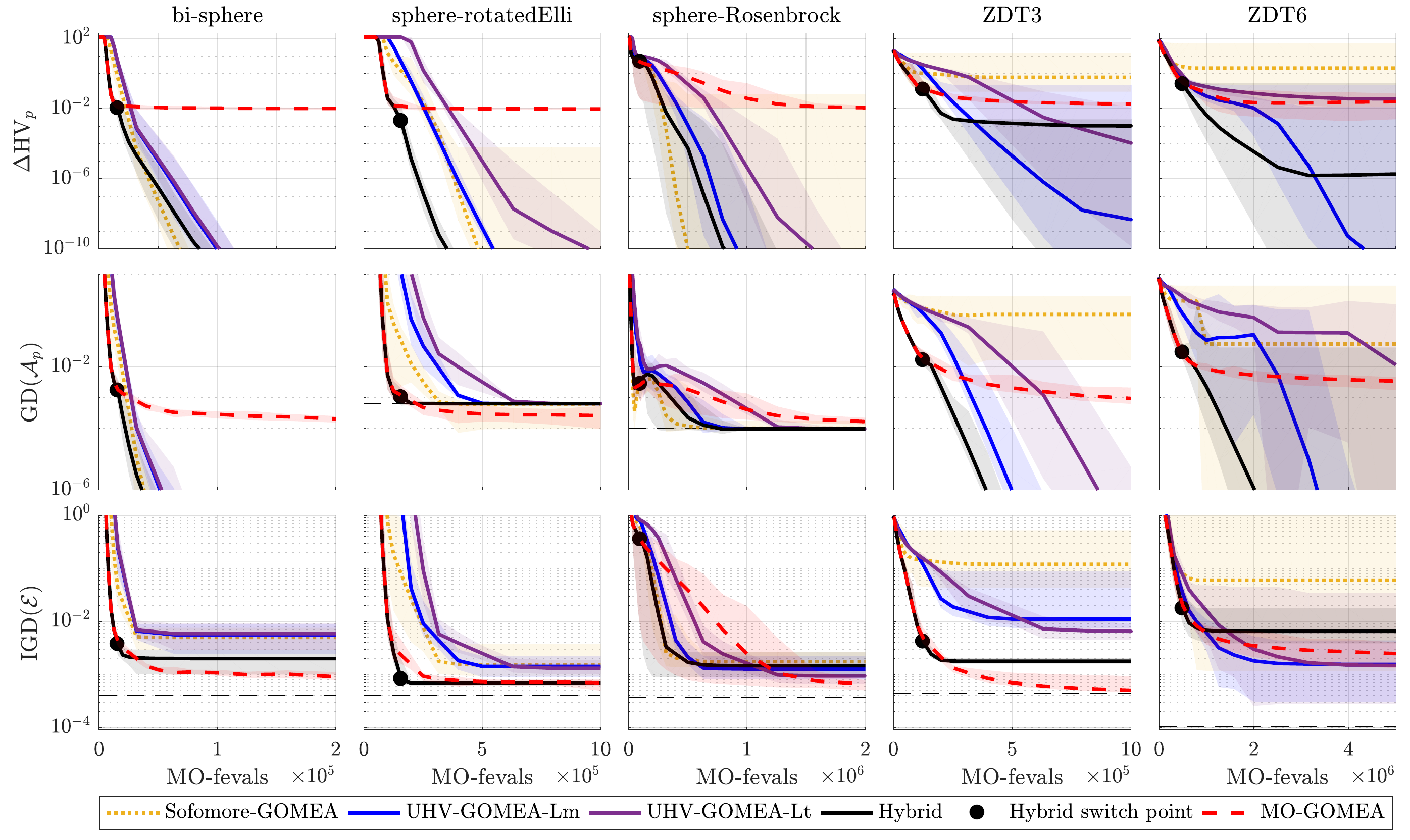}
\caption{Convergence plots for the algorithms discussed in this work. All problems are $n = 10$, $p = 9$, and run with $N=200$, except the bi-sphere problem, which is run with $N=31$. Lines show mean scores; shaded area are min/max scores. Hybrid switch point shows the mean MO-fevals after which MO-GOMEA was terminated and UHV-GOMEA-Lm initialized in the hybrid approach. Dashed black lines shows the maximally achievable GD scores for problems where a finite approximation of the Pareto front is used, and the maximally achievable IGD with a limited archive size $|\cE|\leq1000$ for the IGD.}
\label{fig:emo20_experiment6_hybrid}
\end{figure*}

Convergence results for the discussed algorithms are shown in Figure~\ref{fig:emo20_experiment6_hybrid}. As intended, the hybrid approach terminates MO-GOMEA when stagnation occurs and UHV-GOMEA-Lm takes over from there (indicated by the black dot). The hybrid converges to the Pareto set in all cases, as shown by $\text{GD}(\cA_p)$, and outperformed UHV-GOMEA-Lm, showing that a domination-based initialization is preferable over the uncrowded distance approach. In terms of the archive IGD, the hybrid approach also outperforms UHV-GOMEA-Lm for all problems but ZDT6, where none of the algorithms obtain the maximally achievable IGD.

\subsection{Experiment 6: WFG benchmark}
We benchmark the discussed methods on the commonly used WFG Benchmark \citep{Huband2005}. This test suite consists of 9 benchmark functions with different properties. We consider the instances with $m = 2$ objectives, $k_\text{WFG} = 4$ position variables, and $l_\text{WFG} = 20$ distances variables, resulting in a total of $n = 24$ decision variables. The hypervolume reference point is set to $r = (11,11)$.  Of these problems, WFG1 is separable, but has a flat region in the decision space, which could cause stagnation. WFG2, WFG4, and WFG9 have one or more multimodal objectives. Problems WFG4--9 all have a concave front, WFG1 has a convex front, WFG2 has a disconnected convex front, and WFG3 has a linear front. We solve these benchmark problems with $p = 9$ and a computational budget of $10^7$ MO-fevals. Based on previous results, a population size of $N = 200$ was used for all algorithms. We include two versions of the hybrid method in this experiment. Both first run MO-GOMEA, and Hybrid-Lm then switches to UHV-GOMEA-Lm, while Hybrid-Lt switches to UHV-GOMEA-Lt. All experiments are repeated 30 times. Differences are tested for statistical significance (up to 4 decimals) by a Wilcoxon rank-sum test with $\alpha = 0.05$, pairwise to the best. Ranks (in brackets) are computed based on the mean hypervolume values. All statistics are computed per table. 

\subsubsection{Results}
Results on the WFG benchmark are shown in Table~\ref{tab:emo20_wfg_p9}. Problem WFG1, which has a plateau in its fitness landscape, is consistently solved better with MO-GOMEA. MO-GOMEA als outperforms the other methods on WFG2, which is multimodal, and has a discontinuous front, which prevents the hypervolume-based methods from obtaining the optimal distribution of solutions, as was shown before for the ZDT3 and ZDT6 problems. For the problems with a concave front, WFG4--9, the optimal hypervolume value is $\text{HV}_9^\star = 114.40\ldots$ , which was obtained by many of the algorithms. Especially WFG3 which has optimal hypervolume $\text{HV}_9^\star = 116.50\ldots$, and WFG6 and WFG7 seem to be relatively easy, with none of the methods performing particularly worse than the others. Differences occur often at more than two decimals of accuracy, and are therefore not visible in the table. 

This experiment setup is unfavorable for MO-GOMEA, as it is not aimed to obtain the optimal distribution of exactly $p = 9$ solutions along the front. Despite this bias, it is still very competitive, obtaining the best scores in three of the problems. Especially the difference between UHV-GOMEA-Lm and UHV-GOMEA-Lt is noteworthy, demonstrating that taking linkage into account greatly improves performance on these more difficult problems. Overall, Hybrid-Lt performs best, although differences between different algorithms are small.

\begin{table}
\smaller
\caption{Results on the WFG Benchmark with $p = 9$ MO-solutions, resulting in an $n = 216$ dimensional optimization problem. A computational budget of $10^7$ MO-fevals was used. Hypervolume values $\text{HV}_p$ are shown (mean, $\pm$ standard deviation (rank)), computed with reference point $r = [11,11]$. gHSS was used for MO-GOMEA. Bold numbers are best scores per problems, or those not statistically different from it. Bottom row shows the mean rank, together with the overall rank in brackets.}
\label{tab:emo20_wfg_p9}
\vspace{5pt}
\begin{tabular*}{\textwidth}{@{\extracolsep{\fill}}c@{\extracolsep{\fill}}r@{\extracolsep{\fill}}r@{\extracolsep{\fill}}r@{\extracolsep{\fill}}r@{\extracolsep{\fill}}r@{\extracolsep{\fill}}r@{\extracolsep{\fill}}}
\toprule
\# & Sofomore-GOMEA & UHV-GOMEA-Lm & UHV-GOMEA-Lt & MO-GOMEA & Hybrid-Lm & Hybrid-Lt \\ 
 \midrule 
1  & $96.94{\:\pm\:1.74}$ (5)  & $93.98{\:\pm\:0.88}$ (6)  & $99.22{\:\pm\:2.43}$ (2)  & $\textbf{103.52}{\:\pm\:}1.72$ (1)  & $98.00{\:\pm\:0.88}$ (3)  & $98.00{\:\pm\:0.88}$ (4) \\ 
2  & $110.16{\:\pm\:0.03}$ (5)  & $110.14{\:\pm\:0.01}$ (6)  & $110.21{\:\pm\:0.40}$ (2)  & $\textbf{112.28}{\:\pm\:}3.32$ (1)  & $110.18{\:\pm\:0.00}$ (4)  & $110.18{\:\pm\:0.00}$ (3) \\ 
3  & $116.50{\:\pm\:0.00}$ (3)  & $\textbf{116.50}{\:\pm\:}0.01$ (5)  & $\textbf{116.50}{\:\pm\:}0.00$ (1)  & $116.34{\:\pm\:0.04}$ (6)  & $\textbf{116.50}{\:\pm\:}0.00$ (4)  & $\textbf{116.50}{\:\pm\:}0.00$ (2) \\ 
4  & $112.81{\:\pm\:0.60}$ (5)  & $112.66{\:\pm\:0.64}$ (6)  & $113.23{\:\pm\:0.41}$ (2)  & $113.19{\:\pm\:0.63}$ (3)  & $112.97{\:\pm\:0.56}$ (4)  & $\textbf{113.50}{\:\pm\:}0.55$ (1) \\ 
5  & $\textbf{112.27}{\:\pm\:}0.13$ (1)  & $\textbf{112.08}{\:\pm\:}0.38$ (6)  & $\textbf{112.22}{\:\pm\:}0.00$ (2)  & $112.17{\:\pm\:0.04}$ (4)  & $\textbf{112.14}{\:\pm\:}0.28$ (5)  & $\textbf{112.22}{\:\pm\:}0.00$ (3) \\ 
6  & $\textbf{114.40}{\:\pm\:}0.00$ (1)  & $114.39{\:\pm\:0.02}$ (4)  & $114.39{\:\pm\:0.02}$ (3)  & $114.17{\:\pm\:0.08}$ (6)  & $114.29{\:\pm\:0.38}$ (5)  & $114.39{\:\pm\:0.01}$ (2) \\ 
7  & $\textbf{114.40}{\:\pm\:}0.00$ (3)  & $\textbf{114.40}{\:\pm\:}0.00$ (2)  & $\textbf{114.40}{\:\pm\:}0.00$ (5)  & $114.26{\:\pm\:0.04}$ (6)  & $\textbf{114.40}{\:\pm\:}0.00$ (4)  & $\textbf{114.40}{\:\pm\:}0.00$ (1) \\ 
8  & $111.74{\:\pm\:0.20}$ (3)  & $111.47{\:\pm\:0.29}$ (5)  & $111.51{\:\pm\:0.23}$ (4)  & $111.11{\:\pm\:0.15}$ (6)  & $111.80{\:\pm\:0.04}$ (2)  & $\textbf{111.83}{\:\pm\:}0.01$ (1) \\ 
9  & $111.44{\:\pm\:0.20}$ (6)  & $111.49{\:\pm\:0.05}$ (4)  & $111.48{\:\pm\:0.04}$ (5)  & $\textbf{112.09}{\:\pm\:}0.57$ (1)  & $111.57{\:\pm\:0.25}$ (2)  & $111.57{\:\pm\:0.25}$ (3) \\ 
\midrule & 3.56 (3)& 4.89 (6)& 2.89 (2)& 3.78 (5)& 3.67 (4)& \textbf{2.22 (1)}\\ 
\bottomrule
\end{tabular*}
\end{table}

\section{Discussion}
\label{sec:emo20_discussion}
We formulated the uncrowded hypervolume (UHV) measure, which was used to achieve population-based hypervolume-driven MO optimization using a single-objective problem formulation. We compared this problem formulation to the dynamic interleaved Sofomore framework, which is also hypervolume-based, and the MO problem formulation based on Pareto-dominance that is typically used in MOEAs. These three problem formulations were all solved with versions of the gene-pool optimal mixing evolutionary algorithm (GOMEA) \citep{bouter17,bouter17b}, for a modern and fair comparison. 

We showed that the hypervolume-based methods do not exhibit the stagnation (e.g., in GD) that occurs with domination-based MOEAs, and thereby confirm the results obtained in e.g. \cite{Toure19}. This clearly shows the superiority of hypervolume-based methods when a small number of high-quality solutions is required. However, domination-based MOEAs initially outperform the hypervolume-based methods, especially when the initial population is far away from the Pareto set. A simple hybrid approach, in which a reasonably good approximation set that is obtained with a domination-based MOEAs is used as the initial population of a hypervolume-based algorithm, showed to improve performance compared to the use of both approaches separately. When a large approximation set is required, the difficulty of the hypervolume-based problems increases, and dependency modeling becomes beneficial or even essential for solving them. Additionally, in our experiments, the domination-based MOEA almost always achieves a better elitist archive containing a large number of non-dominated solutions (e.g., $|\cE| = 1000$, in this work). 

MO-GOMEA naturally has diversity-enhancing mechanisms, which might be beneficial for successfully optimizing multimodal problems. The single-objective GOMEA, used here to optimize the hypervolume-based problem formulation, was not particularly developed for this objective function. A better understanding of multi-objective fitness landscapes, of which a first attempt was made in \citep{Kerschke17}, might be helpful to adapt GOMEA, or any other single-objective optimizer, for this specific optimization task.

A limitation of the hypervolume-based approach is that a reference point is required, for which a suitable choice could be unknown in a black-box setting. Additionally, the computational complexity of the hypervolume increases when the number of objectives increases. This makes the UHV expensive for MO optimization problems with $m \geq 3$, although the population size is this application is small, and approximation methods could be used \citep{Bader11,Fieldsend19}. On the other hand, our IBMOP formulation allows MO problems to be solved with a single-objective optimizer, which provides opportunities to explore techniques such as MO multimodal optimization \citep{maree19,tanabe18}, that are well explored for single-objective optimization, but are still upcoming in MO optimization. 

\section{Conclusion}
\label{sec:emo20_conclusion}
We introduced a single-objective problem formulation for multi-objective optimization based on the uncrowded hypervolume (UHV). We showed that problems formulated as such can be efficiently solved with GOMEA by exploiting grey-box properties of this problem formulation. We compared the resulting approach with a version of GOMEA that is based on a classical domination-based selection (MO-GOMEA) and a version that is based on hypervolume optimization (Sofomore-GOMEA). We showed that hypervolume-based optimization can overcome the stagnation from which domination-based methods suffer after a while, and that these methods show convergence to the optimal hypervolume, and thereby to a subset of the Pareto set. However, when the multi-objective problem at hand has difficult landscape features such as multimodality or deceptiveness, the domination-based MO-GOMEA outperformed the hypervolume based methods.

When the desired approximation set size is small, hypervolume-based methods are generally preferable. When the desired approximation set size is large, domination-based methods obtain a better approximation faster. Additionally, in the latter case, the resulting single-objective optimization problem becomes difficult, and dependency modeling becomes essential to still be able to solve the hypervolume-problem up to high accuracy. Hybrid methods, such as the  one proposed in this article, stand the best chance at achieving the overall best performance and being most generally applicable, which also provides a promising area of future research. 

\section*{Acknowledgments} 
\small
This work is part of the research programme IPPSI-TA with project number 628.006.003, which is financed by the Dutch Research Council (NWO) and Elekta. We acknowledge financial support of the Nijbakker-Morra Foundation for a high-performance computing system.

\small

\bibliographystyle{apalike}
\bibliography{Maree_2020}

\end{document}

%% file: math_shortcuts.tex
\newcommand{\powerset}{\raisebox{.15\baselineskip}{\Large\ensuremath{\wp}}}

\DeclareMathOperator{\inX}{\in\cX}

\usepackage{mathrsfs}

\newcommand{\bbN}{\mathbb{N}}

\newcommand{\bbR}{\mathbb{R}}

\newcommand{\be}{\mathbf{e}}

\newcommand{\bff}{\mathbf{f}}

\newcommand{\bm}{\mathbf{m}}

\newcommand{\bx}{\mathbf{x}}

\newcommand{\by}{\mathbf{y}}

\newcommand{\bz}{\mathbf{z}}

\newcommand{\cA}{\mathcal{A}}

\newcommand{\cE}{\mathcal{E}}

\newcommand{\cO}{\mathcal{O}}

\newcommand{\cS}{\mathcal{S}}

\newcommand{\cX}{\mathcal{X}}

\newcommand{\cZ}{\mathcal{Z}}

\newcommand{\abs}[1]{{\left| #1 \right|}} 
\newcommand{\norm}[1]{\lVert #1\rVert} 

\DeclareMathOperator{\hypv}{\text{HV}}
\DeclareMathOperator{\hvi}{\text{HVI}}

\makeatletter
\def\Ddots{\mathinner{\mkern1mu\raise\p@
\vbox{\kern7\p@\hbox{.}}\mkern2mu
\raise4\p@\hbox{.}\mkern2mu\raise7\p@\hbox{.}\mkern1mu}}
\makeatother